\documentclass{article} 
\usepackage{collas2024_conference,times}
\usepackage{easyReview}

\usepackage{amsmath,amsfonts,bm}

\def\eqref#1{equation~\ref{#1}}

\def\1{\bm{1}}

\DeclareMathAlphabet{\mathsfit}{\encodingdefault}{\sfdefault}{m}{sl}
\SetMathAlphabet{\mathsfit}{bold}{\encodingdefault}{\sfdefault}{bx}{n}

\usepackage{hyperref}
\hypersetup{
    colorlinks=true,
    linkcolor=red,
    filecolor=magenta,
    urlcolor=blue,
    citecolor=purple,
    pdftitle={Overleaf Example},
    pdfpagemode=FullScreen,
    }

\title{Formatting Instructions for CoLLAs 2024 \\ Conference Submissions}
\collasfinalcopy

\author{Amir El-Ghoussani  \\
Friedrich-Alexander-Universität Erlangen-Nürnberg\\
\texttt{amir.el-ghoussani@fau.de} 
\And 
Julia Hornauer\\
Ulm University\\
\texttt{julia.hornauer@uni-ulm.de} \\
\AND 
Gustavo Carneiro \\
University of Surrey\\
\texttt{g.carneiro@surrey.ac.uk}
\And
Vasileios Belagiannis\\
Friedrich-Alexander-Universität Erlangen-Nürnberg\\
\texttt{vasileios.belagiannis@fau.de}
}

\newcommand{\etal}{et al.}

\begin{document}

\title{Consistency Regularisation for Unsupervised Domain Adaptation in Monocular Depth Estimation}

\maketitle

\begin{abstract}
In monocular depth estimation, unsupervised domain adaptation has recently been explored to relax the dependence on large annotated image-based depth datasets. However, this comes at the cost of training multiple models or requiring complex training protocols. We formulate unsupervised domain adaptation for monocular depth estimation as a consistency-based semi-supervised learning problem by assuming access only to the source domain ground truth labels. To this end, we introduce a pairwise loss function that regularises predictions on the source domain while enforcing perturbation consistency across multiple augmented views of the unlabelled target samples. Importantly, our approach is simple and effective, requiring only training of a single model in contrast to the prior work. In our experiments, we rely on the standard depth estimation benchmarks KITTI and NYUv2 to demonstrate state-of-the-art results compared to related approaches. Furthermore, we analyse the simplicity and effectiveness of our approach in a series of ablation studies. The code is available at \url{https://github.com/AmirMaEl/SemiSupMDE}.
\end{abstract}

\section{Introduction}
\label{sec:intro}
Monocular depth estimation plays an active role in computer vision with many applications such as 3D pose estimation, semantic segmentation, and 3D reconstruction. Obtaining accurate depth predictions from a single image, though, is an ill-posed problem, especially in complex indoor and outdoor scenes, due to the ambiguous nature of the inverse depth problem. While the standard supervised learning paradigm, using large annotated depth datasets, have achieved promising performance~\citep{eigen_depth_2014,laina2016deeper,agarwal2023attention}, collecting such data requires extensive resources~\citep{make3dpaper,gaidon_vkitti_2016,handa_scenenet_2015}.
\begin{figure}[h]
  \centering
   \includegraphics[width=0.98\linewidth]{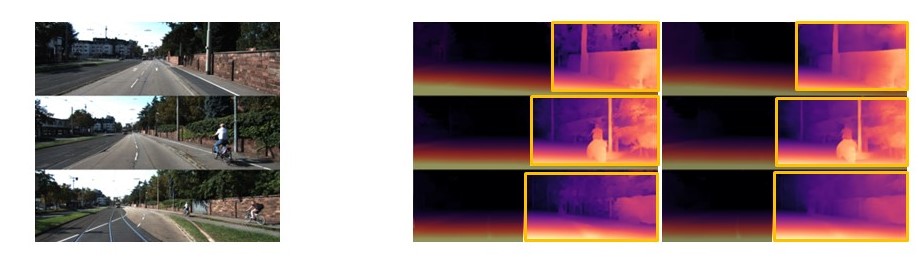}
   \put(-425,-2){Input KITTI images.}
   \put(-250,-2){Predictions after pretraining}\put(-200,-12){ in $D_S$.}
   \put(-120,-2){Predictions after proposed}\put(-120,-12){ consistency-based training.}
   \caption{
   Following pretraining on the source domain using CutMix~\citep{yun_cutmix_2019} data augmentation, we observe that the resulting depth predictions show some fidelity but could benefit from further refinement in certain areas. Specifically, they appear overly fragmented or "edgy" in localised regions (highlighted in the figure). Based on this observation we carefully design our consistency-based approach for domain adaptation in monocular depth estimation, to particularly smooth localised and fragmented regions.}
    \label{fig:teaser}
\end{figure}
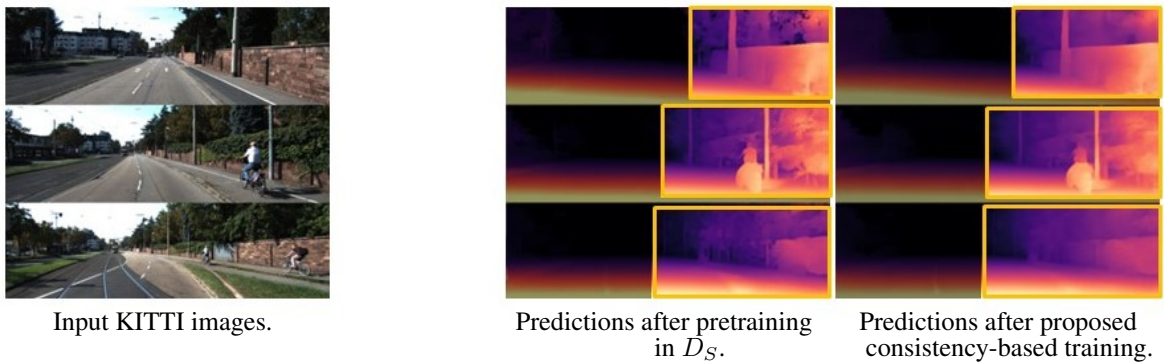

To relax the data annotation requirement, unsupervised learning approaches exploit geometric constraints between multiple views of a scene. For example, stereo-based methods leverage left-right consistency between rectified stereo image pairs to train without explicit depth supervision~\citep{godard2017unsupervised,park2022unsupervised}. Such methods require calibrated stereo cameras during both training and inference. Another line of work utilises temporal consistency in video sequences to derive implicit depth cues from camera motion~\citep{zhou2017unsupervised,mahjourian2018unsupervised,zou2018df}. While these approaches do not require labelled training data, they are limited to scenarios where stereo imagery or video sequences are available.
Furthermore, unsupervised learning methods struggle in situations involving occlusion and object motion~\citep{wang2018occlusion,sun2023unsupervised}.
In contrast, semi-supervised learning methods utilise a subset of labelled images to directly supervise the model on the depth prediction task~\citep{kuznietsov2017semi, amiri2019semisupervised}. This guidance is stronger than the indirect cues from multi-view constraints alone in unsupervised learning.
More importantly, semi-supervised approaches can leverage unlabelled data along with a limited amount of labelled data, even if stereo or video data is scarce or unavailable in the target domain. However, semi-supervised monocular depth estimation techniques often rely on labelled and unlabelled samples being drawn from the same underlying distribution~\citep{kuznietsov2017semi}. 
In practical domain adaptation scenarios, this assumption may not always hold since the labelled source and unlabelled target domains exhibit distribution shifts~\citep{akada2022self,lopez-rodriguez_desc_2020}.

To address this issue, recent approaches in monocular depth estimation have explored unsupervised domain adaptation which leverages available labelled source data together with unlabelled target domain images~\citep{yen_3d-pl_2022,kundu_adadepth_2018,zheng_t2net_2018,gasda_2019,lopez-rodriguez_desc_2020}. 
 
In particular, unsupervised domain adaptation techniques leverage labelled source data to guide prediction in unlabelled target domains by aligning representations across domains~\citep{qiu2016unrealcv,shrivastava2017learning}. These are the key capabilities that we also aim to explore in this work.
Existing domain adaptation approaches for monocular depth estimation mainly rely on self-supervised or unsupervised techniques using consistency-based~\citep{zheng_t2net_2018,xie2016deep3d,zhou2017unsupervised} or geometry-based~\citep{yen_3d-pl_2022,yin2018geonet} losses. While consistency methods provide a general constraint using simple transformations, geometry models require additional metadata, e.g., camera intrinsics, that may be unavailable for new unlabelled target domains. Similarly, our approach also falls into the first category of consistency-based approaches. {In contrast to prior consistency-based works that focused only on target-domain consistency regularisation, we incorporate supervised information by matching predictions on the target and source images simultaneously. Additionally, we employ consistency regularisation for the purpose of unsupervised domain adaptation by requiring target predictions to remain coherent when propagated through multiple augmentation streams. To our knowledge this represents the first application of a technique using both multi-stream target consistency and source-based regularisation for domain adaptive monocular depth estimation. }

Furthermore, the existing domain adaptation approaches often involve complex multi-stage optimising various geometric and photometric losses modelling camera motion or stereo correspondences~\citep{zou2018df,mahjourian2018unsupervised}, leading to optimisation challenges. For example, external models such as semantic segmentation models are used to enforce semantic consistency~\citep{lopez-rodriguez_desc_2020} or 3D point cloud models~\citep{yen_3d-pl_2022} are employed for pseudo label generation, ultimately increasing dependencies.
In contrast, we present an approach to train only the depth estimation model and require no additional trainable models, keeping the training procedure simple.
Additionally, target and source data are typically optimised separately for the above methods without an explicit linking of predictions across domains.
In comparison, we propose to simultaneously optimise the supervised source objective and the unsupervised target objective in a single training stage.

In this work, we formulate the problem of unsupervised domain adaptation for monocular depth estimation as a consistency-based semi-supervised learning problem. We present a semi-supervised approach for training a single model by assuming access only to the labels of the source domain data. Initially, we conduct supervised pre-training using heavily augmented labelled source data to obtain preliminary depth predictions. Afterwards, depth maps predicted on unlabelled target domain data are used as pseudo-labels. Our approach is motivated by the fact that after the initial pre-training in the source domain, the depth predictions appear somewhat faithful but are still edgy and fragmented in certain areas (see Fig.~\ref{fig:teaser}). This suggests that additional refinement of object boundaries and detection of previously missing objects could improve the overall depth map quality. To that end, we introduce a pairwise loss term that regularises the predictions on the source domain to be consistent with the ground truth depth distribution in the source domain, while simultaneously enforcing multiple perturbation consistency during pseudo-labelling on the target domain. We assess the performance of our method in two standard domain adaptation benchmarks with indoor and outdoor datasets, where Virtual KITTI~\citep{gaidon_vkitti_2016} / SceneNet~\citep{handa_scenenet_2015} are the source domains and KITTI~\citep{geiger_vision_2013_kitti} / NYU~\citep{nyu_depth} are the corresponding target domains. 

Our main contributions can be summarised as follows:
\vspace{-\topsep}
\begin{itemize}
\setlength{\topsep}{0pt}
    \setlength{\itemsep}{0pt}%
    \setlength{\parskip}{0pt}
    \item We present a semi-supervised approach for unsupervised domain adaptation by introducing a pairwise loss that regularises the predictions on the source
domain, while simultaneously enforcing perturbation consistency across multiple augmented views of unlabelled target data.
   \item Importantly, we do not require any additional trainable models, making the proposed method simple and effective. To the best of our knowledge, we are the first to develop domain adaptation by simultaneous optimisation using source and target data in one training stage.
   \item In the extensive evaluation based on the standard depth estimation benchmarks KITTI~\citep{geiger_vision_2013_kitti} and NYU~\citep{nyu_depth}, our proposed method demonstrates state-of-the-art performance in the domain adaptation task.
\end{itemize}

\section{Related Work}

\subsection{Monocular Depth Estimation}
Monocular depth estimation consists of predicting the depth information of a scene from a single image. 
With the rapid development of deep neural networks, monocular depth estimation with the help of deep learning has been widely studied. 
Various supervised learning approaches have been proposed in recent years for monocular depth estimation. Eigen~\etal~\citep{eigen_depth_2014} propose a CNN-based approach to directly regress the depth. 
Liu \etal~\citep{liuLearningDepthSingle2016} propose utilising a fully-connected CRF as a post-processing step to refine monocular depth predictions. Other methods have extended the CNN-based network by changing the regression loss to a classification loss~\citep{caoEstimatingDepthMonocular2016,fuDeepOrdinalRegression2018,pengwangUnifiedDepthSemantic2015,yinEnforcingGeometricConstraints2019,bhat2021adabins} or change the architecture of the depth estimation network entirely~\citep{liu2023vadepthnet,rudolph2022lightweight}.
In addition, a number of self-supervised techniques have been proposed to train monocular depth estimation models without relying on ground truth labels.  GeoNet~\citep{yin2018geonet} enforces epipolar geometry consistency, and DF-Net~\citep{zou2018df} models rigid scene flow. SfMLearner~\citep{zhou2017unsupervised} explores photometric alignment. \citet{godard2017unsupervised} use stereo consistency between synchronised camera views for unsupervised monocular depth estimation. 
More recently, attention-based architectures have achieved new performance milestones in monocular depth estimation. DepthFormer adopted Vision Transformers for strong multi-scale representations~\citep{lin2020space}. Similarly, DPT fused CNNs and self-attention in a hybrid network~\citep{ranftl2021vision}. Neural  Window Fully Connected CRFs~\citep{yuan2022new} introduced CRF modelling in a neural window, leveraging pixel-level affinity and semantic connections to refine initial depth maps. \citet{agarwal2023attention} propose a novel architecture that utilises skip attention to refine pixel-level queries and improve depth predictions.
While supervised learning approaches have achieved considerable success in monocular depth estimation tasks, acquiring large annotated depth datasets at scale remains challenging.
Self-supervised methods provide an alternative, but traditional techniques depend on camera parameters and frame geometries, which often do not hold strictly in real-world settings. Furthermore, the lack of direct supervision signals hinders the ability of self-supervised models to fully leverage available visual cues.

\subsection{Domain adaptation for depth estimation}

Domain adaptation techniques leverage synthetic data as a labelled source domain and unlabelled real data as the target domain for monocular depth estimation~\citep{chen2019crdoco,kundu_adadepth_2018,lopez-rodriguez_desc_2020,pnvr_sharingan_2020,gasda_2019,zheng_t2net_2018,yen_3d-pl_2022}. Most prior approaches treat depth estimation as a regression task while relying on style- or image translation for pixel-level adaptation~\citep{atapour}, adversarial learning for feature-level adaptation~\citep{kundu_adadepth_2018,visDomAdap}, or a combination of both~\citep{zheng_t2net_2018,gasda_2019}. For instance, AdaDepth~\citep{kundu_adadepth_2018} aligns source and target distributions in features and predictions. $T^2Net$~\citep{zheng_t2net_2018} extends this idea by adopting both synthetic-to-real translation network and feature alignment in the task network. However, aligning features were found to be ineffective in outdoor datasets despite promising improvements observed when training on real stylised images~\citep{zheng_t2net_2018}. 3D-PL~\citep{yen_3d-pl_2022} combines domain adaptation, pseudo-labelling and 3D-awareness to improve the depth estimation performance in a target domain. These methods leverage the geometric information present in depth maps to generate more accurate pseudo-labels. Other follow-up methods have explored bidirectional translation (real-to-synthetic and synthetic-to-real) and utilise depth consistency loss on the predictions between real and real-to-synthetic images~\citep{gasda_2019,chen2021s2r}. GASDA~\citep{gasda_2019} incorporates additional information: it utilises stereo pairs to encourage geometry consistency and align stereo images for depth estimation. Similarly, SharinGAN \citep{pnvr_sharingan_2020} maps domains to a shared space with geometry constraints. Moreover, DESC \citep{lopez-rodriguez_desc_2020} adopts segmentation for pseudo-labeling based on instance properties.
Our approach to domain adaptation for depth estimation offers several advantages compared to existing methods. Specifically, our method provides direct pseudo-labels on real data in a simple and straightforward manner, without the need for any external information, such as semantic information~\citep{lopez-rodriguez_desc_2020}. 
{Additionally we leverage labeled source data by simultaneously optimizing on target and source domain images. }

\subsection{Semi-supervised learning}
There are two main methodologies used in semi-supervised learning, namely entropy minimisation and consistency regularisation. Entropy minimisation involves self-training on pseudo labels for unlabelled data. Conversely, consistency regularisation assumes predictions should remain stable under perturbations. Among the consistency regularisation methods, FixMatch~\citep{sohn_fixmatch_2020} proposes using strong perturbations for unlabelled images and supervising the training process with predictions from weakly perturbed images, combining the benefits of both approaches. Recent advancements like FlexMatch~\citep{zhang2021flexmatch} and FreeMatch~\citep{wang2022freematch} consider class-wise confidence thresholds. In \cite{yang2023revisiting}, UniMatch is proposed, which aims to improve the performance of semi-supervised semantic segmentation by incorporating extra information and imposing constraints on real images.
Our work takes inspiration from these semi-supervised learning techniques that leverage consistency regularisation principles. However, these approaches have primarily focused on classification and segmentation tasks with discrete label spaces. As monocular depth estimation produces continuous-valued depth maps, it poses a regression problem rather than classification or segmentation.
Inspired by FixMatch~\citep{sohn_fixmatch_2020}, we propose a framework for semi-supervised monocular depth estimation. Specifically, we formulate depth prediction as regression and design augmentations and consistency losses suitable for continuous outputs.

\section{Method}
Let $f(\mathbf{x};\theta)$ denote a neural network parameterised by $\theta$ that maps the input image $\mathbf{x}\in \mathbb{R}^{h \times w \times 3}$ to the depth prediction $\hat{\mathbf{y}} \in \mathbb{R}^{ h \times w\times 1}$. The depth estimation model $f(\cdot)$ is trained on the source domain $D_S$ containing images $\mathbf{x}_s$ with corresponding ground truth depth maps $\mathbf{y}_s$, and the target domain $D_T$ of images $\mathbf{x}_u$ without accompanying depth labels.
Directly optimising $f(\cdot)$ based solely on supervised losses in the source domain $D_S$ would lead to degraded performance when applied to the unlabelled target domain $D_T$, owing to the domain gap between the source and the target data distributions. 

To address this issue, we propose a semi-supervised learning approach that facilitates direct supervision on the target image $\mathbf{x}_u$, thereby effectively reducing the domain gap. Initially, we train the depth prediction model $f(\cdot)$ using  source domain data. After this pre-training, we train our model by enforcing perturbation consistency across augmented views in $D_T$ while also optimising on supervised source data using a new pairwise loss. Critically, we simultaneously apply both loss terms during training, allowing source supervision to guide pseudo-label refinement while the proposed consistency across target perturbations induces adaptation. 


\begin{figure*}[h]
  \centering
     \includegraphics[width=0.99\linewidth]{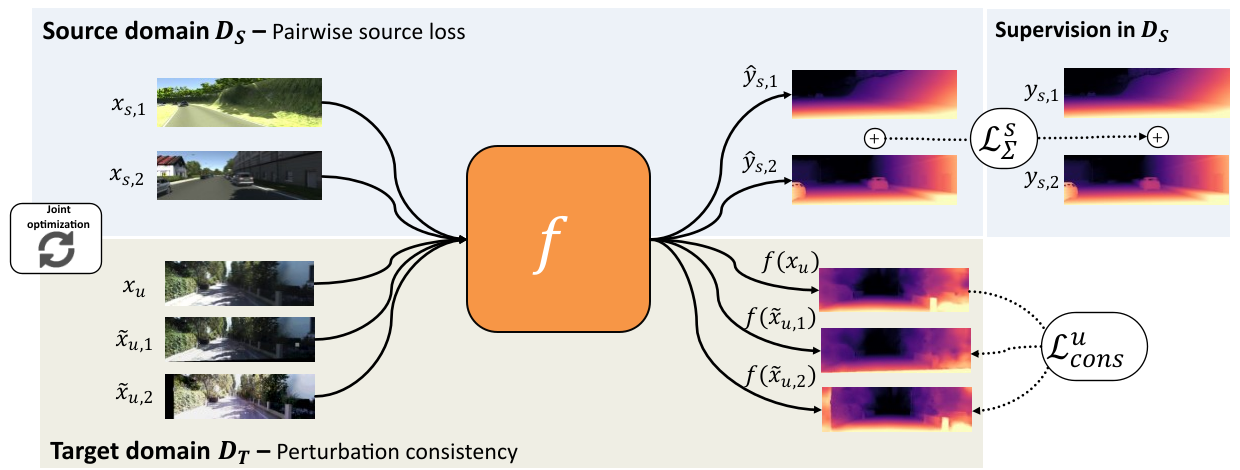}
     \caption{Overview of the approach. Initially, we sample two independent source domain images $\mathbf{x}_{s,1}$ and $\mathbf{x}_{s,2}$ along with one target domain image $\mathbf{x}_u$. This target domain image is then fed into two perturbation streams, in which independent augmentations are applied to the target image, denoted as $\Tilde{\mathbf{x}}_{u,1}$ and $\Tilde{\mathbf{x}}_{u,2}$. In total five samples are concatenated and fed into the depth estimation model $f(\cdot)$. Afterwards, the predictions are chunked back into their initial shapes. The loss on the supervised domain is computed by enforcing consistency between the sum of predictions and the sum of the two ground truth samples. Finally, the unsupervised loss is calculated by enforcing consistency between generated perturbations. Yellow and blue colors correspond to the supervised source and the unsupervised target domain, respectively.}
   \label{fig:onecol}
\end{figure*}
\paragraph{Preliminaries}
\label{sec:preliminaries}
We derive inspiration by FixMatch~\citep{sohn_fixmatch_2020} in which each unlabelled sample undergoes simultaneous perturbations by two operators: a weak perturbation such as cropping and a strong perturbation like colour jittering. Weak perturbations are generally augmentations, that minimally alter the scene content, while strong perturbations introduce more distortion and artefacts.
The overall objective function is a combination of supervised loss $\mathcal{L}^s$ and unsupervised loss $\mathcal{L}^u$, given by:
\begin{equation}\label{FixMatchLoss}
\mathcal{L} = \frac{1}{2}(\mathcal{L}^s + \mathcal{L}^u)
\end{equation}
Typically the supervised term $\mathcal{L}_s$ is a cross-entropy loss between the model predictions and the labels.
The unsupervised loss $\mathcal{L}_u$ serves to regularise the prediction of the sample under strong perturbations to be the same as that under weak perturbations.

Based on Eq.~\ref{FixMatchLoss}, we present a formulation for the monocular depth estimation setting. 

We propose to replace the supervised loss $\mathcal{L}^s$ with a pairwise loss $\mathcal{L}_{\Sigma}^s$, applied to source domain samples in $D_S$, while also extending $\mathcal{L}^u$ with perturbation consistency across three augmented views in the target domain $D_T$.

\subsection{Perturbation consistency}

We seek to minimise the differences between predictions on augmented versions of the same input image. Due to the unsupervised domain adaptation setting where no depth annotation for the target domain is available, we rely on enforcing consistency between multi-stream perturbed image predictions in the target domain. For each input sample, we apply the RandAugment~\citep{cubuk_randaugment_2019} algorithm to generate multiple perturbed views, or "streams", of the data. Each augmentation chain is applied independently to a target sample $\mathbf{x}_u$ to produce uniquely transformed versions of the input. In total, we generate two augmented versions of the image, in addition to the initial unlabelled target domain image $\mathbf{x}_u$:
\begin{equation}
\tilde{\mathbf{x}}_{u,1} = RA_{Depth,1}(\mathbf{x}_u),
\end{equation}
\begin{equation}
\tilde{\mathbf{x}}_{u,2} = RA_{Depth,2}(\mathbf{x}_u),
\end{equation}
where $RA_{Depth,1}$ and $RA_{Depth,2}$ represent independent RandAugment operations. 
The multi-stream perturbed images $\tilde{\mathbf{x}}_{u,1}$ and $\tilde{\mathbf{x}}_{u,2}$ along with an initial sample $\mathbf{x}_u$ are then fed through the depth estimation model $f(\cdot)$ to obtain three predicted depth maps. All three depth predictions should be consistent; therefore, we compute the $\mathcal{L}_1$ loss between the predicted depth maps to enforce consistency:
\begin{align*}
\mathcal{L}_{cons}^u = \sum_{i=1}^N \lVert f(\mathbf{x}_{u,i}) - f(\tilde{\mathbf{x}}_{u,1,i})\rVert_1 + \lVert f(\mathbf{x}_{u,i})- f(\tilde{\mathbf{x}}_{u,2,i})\rVert_1,
\end{align*}
where $N$ is the batch size. We empirically observed that two augmented versions of each unlabelled sample provided the best trade-off between boosting performance via view redundancy and managing memory costs. Moreover, prior work on semi-supervised learning has shown that generating multiple perturbed views of unlabelled data strengthens consistency regularisation~\citep{berthelot2019remixmatch,caron2020unsupervised}.

\subsection{Pairwise loss}

While the unsupervised perturbation consistency adapts the model to the target domain without source supervision, there is no way to directly assess the quality and accuracy of pseudo-labels $f(\mathbf{x}_u), f(\Tilde{\mathbf{x}}_{u,1})$ and $f(\Tilde{\mathbf{x}}_{u,2})$. Errors can accumulate during training as the network learns potentially incorrect target predictions. For that reason, we regularise the training process, leveraging a pairwise loss $\mathcal{L}^s_{\Sigma}$ in the source domain $D_S$. Specifically, for each source domain batch we sample image pairs $(\mathbf{x}_{s,1},\mathbf{x}_{s,2})$ along with their corresponding depth annotations $(\mathbf{y}_{s,1},\mathbf{y}_{s,2})$, and then compute the $\mathcal{L}_1$ distance between the sum of predictions and labels over a batch of size $N$:
\begin{equation}
\hat{\mathbf{y}}_{s,1} = f(\mathbf{x}_{s,1}),\quad  \hat{\mathbf{y}}_{s,2} = f(\mathbf{x}_{s,2}) 
\end{equation}
\begin{equation}
\hat{\mathbf{y}}_{\Sigma,s} = \hat{\mathbf{y}}_{s,1} + \hat{\mathbf{y}}_{s,2}
\end{equation}
\begin{equation}
    \mathbf{y}_{\Sigma,s} = \mathbf{y}_{s,1}+\mathbf{y}_{s,2}
\end{equation}
\begin{equation}
    \mathcal{L}_{\Sigma}^s = \sum_{i=1}^N||\mathbf{y}_{\Sigma,s,i} - \hat{\mathbf{y}}_{\Sigma,s,i}||_1.
\end{equation}

A key observation from our experiments is that the proposed pairwise source regularisation approach from the source domain labels requires using an extremely low learning rate for stable optimisation. Since already seen images are combined in new configurations during each summation, essentially, this approach ends up augmenting the training data online.

\subsection{Complete loss function} 

In total, we propose a loss formulation that combines the multiple perturbation consistency constraints imposed on unlabelled target samples with the proposed summation-based pairwise loss computed on the labelled source samples. It is given by:
\begin{equation}
    \mathcal{L}_{total} = \frac{1}{2}\sum_i^{N}({\mathcal{L}^s_{\Sigma,i} + \mathcal{L}^u_{cons,i})\Big|_{r=2}}.
\end{equation}
where $N$ denotes the batch size.
Additionally, we introduce the batch ratio hyperparameter $r=\frac{n_{supervised}}{n_{unsupervised}}$ to balance the amount of supervised vs. unsupervised training data. A ratio parameter $2$ indicates that twice as many supervised as unsupervised samples were used during one forward pass (an exact calculation of the total amount of samples in one forward pass with $N=12$ and $r=2$ is shown in the appendix, in Eq.~\ref{eq:pairlossbs}).
An ablation (see Table~\ref{tab:valuesr}, in Sec. \ref{sec:ablations}) examines various $r$ values, revealing a static value of 2 outperforms other schemes.

\subsection{Model Pretraining}
\label{sec:pretraining}
During model pre-training, images from one single batch are mixed with one another using CutMix~\citep{yun_cutmix_2019} augmentations. 
This prevents the depth estimation model from relying too heavily on local image details, as it forces the model to consider information from different regions. 
We apply CutMix to all source samples $\mathbf{x}_s$ and their corresponding annotations $\mathbf{y}_s$ during pretraining. 
CutMix randomly combines two different images by cutting out a random patch from one image and replacing it with the same-sized patch from another image. The corresponding ground truth labels are also mixed in proportion to the area of the patches, defined by a hyperparameter $\alpha$. Fig. \ref{fig:cutmix} illustrates this procedure for the virtual KITTI source dataset~\citep{gaidon_vkitti_2016}.  We then compute the $\mathcal{L}_1$-distance between depth predictions of these augmented samples $\tilde{\mathbf{x}}_{cm,s}$ and their corresponding ground truth depth annotation $\tilde{\mathbf{y}}_{cm,s}$ over the batch of size $N$:
\begin{figure*}[]
  \centering
    \includegraphics[width=0.98\textwidth]{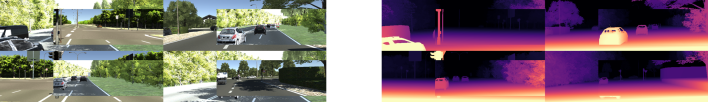}
    \put(-360,-15){$\mathbf{x}_{cm,s}$}
    \put(-110,-15){$\mathbf{y}_{cm,s}$}\\
    \caption{CutMix augmented input images $\mathbf{x}_{cm,s}$ in the source domain along with their corresponding augmented ground truth depth annotation $\mathbf{y}_{cm,s}$. We choose $\alpha = 0.5$, controlling the patch size of the CutMix augmentation.}
    \label{fig:cutmix}
    \end{figure*}
\begin{equation}
    \mathcal{L}^s  = \sum_{i=1}^N ||f(\tilde{\mathbf{x}}_{cm,s,i}) - \tilde{\mathbf{y}}_{cm,s,i}||_1.
\end{equation}

\section{Experiments}
We assess our method on two challenging benchmarks, one indoors and one outdoors. Afterwards, we conduct detailed ablation studies on each benchmark to analyse the importance of the different components in our approach.

\subsection{Datasets}
\paragraph{Outdoor}
We utilise the virtual KITTI (vKITTI)~\citep{gaidon_vkitti_2016} dataset, with 21,260 synthetic image-depth pairs as our labeled source domain, and the real KITTI dataset~\citep{geiger_vision_2013_kitti}, containing 22,600 images as the unlabeled target domain. As vKITTI and KITTI contain outdoor street scenes, we preprocessed both and clip maximum depth values to 80~m to ensure consistency when applying our approach during training, similar to \citep{zheng_t2net_2018}. Following prior works~\citep{zheng_t2net_2018,yen_3d-pl_2022,gasda_2019}, we evaluate the performance on the Eigen test split~\citep{eigen_depth_2014}. For the final test evaluation, we use the Garg crop~\citep{garg2016unsupervised} and clamp the ground truth depth to 50~m and 80~m following prior work~\citep{yen_3d-pl_2022}. 

\paragraph{Indoor}
For our second experiment, we employ the large-scale SceneNet synthetic dataset~\citep{handa_scenenet_2015} as the source and evaluated the real-world NYUv2~\citep{nyu_depth} indoor dataset as the target domain. Additionally, we apply a filtering procedure to the SceneNet dataset to better match the characteristics of NYUv2. Specifically, we aim to resemble the depth distribution of NYUv2, as SceneNet is considerably larger. After filtering, our source SceneNet set contains 4,737 samples. We combine this filtered SceneNet set with a randomly selected subset of 5,000 samples from the raw NYUv2 data as our target domain samples. By selecting a target domain subset of similar size to our source domain set, we aim to keep the ratio of unsupervised to supervised samples close to $1:1$ during training.

Given these datasets capture indoor room layouts and objects, significantly shorter maximum depths are present. Therefore, we clip the maximum depth to 8~m for this indoor adaptation problem, following prior work~\citep{zheng_t2net_2018}.
Our final evaluation is then performed on the Eigen split~\citep{eigen_depth_2014} of the NYU depth data, clipped to 8~m.
\paragraph{Implementation details}
We adopt a  U-Net like~\citep{ronneberger_u-net_2015} architecture similar to \cite{zheng_t2net_2018} as our depth prediction model $f(\cdot)$. The model is trained on an NVIDIA GTX 6000 GPU using the Adam optimizer~\citep{Kingma2014AdamAM}, pretraining is done with a learning rate of $4 * 10^{-3}$ for $250$ epochs while our domain adaptation is applied at learning rate of $4*10^{-8}$ for $10$ epoch with a linear decay after 4 epochs. In total pretraining takes about six hours for the outdoor benchmarks and about one hour for the indoor benchmarks.

\paragraph{Evaluation metrics}
Consistent with previous research~\citep{yen_3d-pl_2022,zheng_t2net_2018,kundu_adadepth_2018,gasda_2019}, we assess unsupervised domain adaptation for monocular depth prediction. To evaluate the adapted depth estimation model, we utilize common metrics, including absolute relative error (AbsRel), root mean squared error (RMSE), squared relative error (SqRel), logarithm of root mean squared error (RMSE Log), and accuracy under different threshold values ($A1$ at threshold $\delta < 1.25$, $A2$ at $\delta < 1.25^2$ and $A3$ at $\delta<1.25^3$).
\paragraph{Comparison  to Related Work}
For a fair evaluation, we compare to prior work using analogous experimental protocols. Specifically, we benchmark against fully-supervised methods training on single domains \citep{eigen_depth_2014, liuLearningDepthSingle2016}. We also compare to baselines trained on separate source/target distributions without adaptation. Critically, we evaluate against prominent unsupervised domain adaptation methods \citep{kundu_adadepth_2018, zheng_t2net_2018, yen_3d-pl_2022} which similarly utilise unlabelled target data through depth completion or translation. These prior works employ architectures and datasets matching our setup, allowing an effective and fair assessment of our approach. 
{The works of SharingGAN~\citep{pnvr_sharingan_2020} and GASDA~\citep{gasda_2019} report results when leveraging stereo imagery under the assumption that epipolar geometry constraints can be imposed. Consequently we do not compare our approaches to theirs.}

\begin{table*}[h]
\centering
\caption{Results on the KITTI Eigen split (80m cap) for single image depth estimation methods. The column 'supervised' determines whether an approach has been trained with target image ground truth samples. 'K' refers to KITTI, 'S' to the vKITTI and 'CS' to Cityscapes data. 'DA' refers to domain adaptation approaches.}
\label{tab:kitti80m}
\footnotesize
\begin{tabular}{cccccccccccccc}
\hline
Method & Supervised & Dataset & AbsRel$\downarrow$ & SqRel$\downarrow$ &RMSE$\downarrow$
 & RMSE log$\downarrow$ & A1$\uparrow$&A2$\uparrow$&A3$\uparrow$ \\
 \hline
 \cite{eigen_depth_2014} & Yes & K & 0.203 & 1.548 & 6.307 & 0.282 & 0.702 & 0.890 & 0.958 \\
\cite{liuLearningDepthSingle2016} & Yes & K  & 0.202 & 1.614 & 6.523 & 0.275 & 0.678 & 0.895 & 0.965 \\
\cite{zhou2017unsupervised} & No & K  & 0.208 & 1.768 & 6.856 & 0.283 & 0.678 & 0.885 & 0.957 \\
\cite{zhou2017unsupervised} & No & K+CS & 0.198 & 1.836 & 6.565 & 0.275 & 0.718 & 0.901 & 0.960 \\
\hline
All synthetic & No & S  & 0.253 & 2.303 & 6.953 & 0.328 & 0.635 & 0.856 & 0.937 \\
All real & No & K  & 0.158 & 1.151 & 5.285 & 0.238 & 0.811 & 0.934 & 0.970 \\
\hline
\cite{kundu_adadepth_2018} & No & K+S(DA)  & 0.214 & 1.932 & 7.157 & 0.295 & 0.665 & 0.882 & 0.950 \\
\cite{zheng_t2net_2018} & No & K+S(DA)  & 0.182 & 1.611 & 6.216 & 0.265 & 0.749 & 0.898 & 0.959 \\
\cite{yen_3d-pl_2022} & No & K+S(DA) & 0.169 & 1.371 & 6.037 & 0.256 & 0.759 & 0.904 & \textbf{0.961} \\
Ours & No & K+S(DA) & \textbf{0.168} & \textbf{1.347} & \textbf{6.002} & \textbf{0.254} & \textbf{0.769} & \textbf{0.907} & \textbf{0.961}\\
\hline
\end{tabular}
\end{table*}

\begin{table*}[ht]
\centering
\caption{Results on the KITTI Eigen split capped to 50m. 'K' refers to KITTI, 'S' to the vKITTI and 'CS' to Cityscapes data. 'DA' refers to domain adaptation approaches. }
\label{tab:kitti50m}
\footnotesize
\begin{tabular}{cccccccccccccc}
\hline
Method & Dataset & AbsRel$\downarrow$ & SqRel$\downarrow$ &RMSE$\downarrow$& RMSE log $\downarrow$& A1$\uparrow$ &A2$\uparrow$&A3$\uparrow$ \\
\hline
\cite{garg2016unsupervised}  & K  & 0.169 & 1.080 & 5.104 & 0.273 & 0.740 & 0.904 & 0.962 \\
\hline
All synthetic &  S  & 0.244 & 1.771 & 5.354 & 0.313 & 0.647 & 0.866 & 0.943 \\
All real &  K & 0.151 & 0.856 & 4.043 & 0.227 & 0.824 & 0.940 & 0.973 \\
\hline
\cite{kundu_adadepth_2018} &  K+S(DA)  & 0.203 & 1.734 & 6.251 & 0.284 & 0.687 & 0.899 & 0.958 \\
\cite{zheng_t2net_2018} & K+S(DA)  & 0.168 & 1.199 & 4.674 & 0.243 & 0.772 & 0.912 & 0.966 \\
\cite{yen_3d-pl_2022} & K+S(DA)  & 0.162 & 1.049 & 4.463 & 0.239 & 0.776 & 0.916 & \textbf{0.968} \\
Ours & K+S(DA) & \textbf{0.161} & \textbf{1.028} & \textbf{4.449} & \textbf{0.237} & \textbf{0.785} & \textbf{0.918} & \textbf{0.968}\\
\hline
\end{tabular}
\end{table*}

\begin{table*}[t]
\centering
\caption{Results of our approach on the NYU Eigen split capped at 8m. 'N' refers to NYU data, 'S' refers to SceneNet data and 'DA' corresponds to domain adaptation.}
\label{tab:scenenet}
\footnotesize
\begin{tabular}{ccccccccccccc}\hline
Method &  Dataset &  AbsRel$\downarrow$ & SqRel$\downarrow$ &RMSE$\downarrow$& RMSE log $\downarrow$& A1$\uparrow$ &A2$\uparrow$&A3$\uparrow$ \\\hline
all-real & N & 0.201 & 0.252 & 0.789 & 0.288 & 0.609 & 0.862 & 0.956\\
all-synthetic & S & 0.432 & 0.720 & 1.183 & 0.439 & 0.395 & 0.686 & 0.859 \\\hline
\cite{zheng_t2net_2018} & N+S(DA)  & \textbf{0.394} & 0.613 & 1.156 & 0.417 & 0.407 & 0.707 & 0.878 \\

Ours & N+S  & 0.417 & \textbf{0.601} & \textbf{1.055} & \textbf{0.385} & \textbf{0.426} & \textbf{0.725} & \textbf{0.890} \\\hline
\end{tabular}
\end{table*}

\subsection{Results}
\paragraph{Outdoor experiments} Table~\ref{tab:kitti80m} shows quantitative results where we compare our approach to various baselines and state-of-the-art approaches with ground truth depth labels clipped to 80~m. In addition, we compare to 'all-synthetic' and 'all-real' baselines, referring to the approaches where the model is trained on synthetic or real data only. These can be seen as the upper and lower bound, respectively. We also provide quantitative results capped to 50~m in Table~\ref{tab:kitti50m}. In this setup, we compare to the supervised approach proposed by Garg et al.~\citep{garg2016unsupervised}. 
In comparison to the standard supervised training baseline 'all-synthetic', our method outperforms this baseline by a significant margin. We further compare to three previous state-of-the-art semi-supervised depth estimation techniques. Notably, our approach achieves a significant improvement in $A1$-accuracy compared to prior work and outperforms 3D-PL~\citep{yen_3d-pl_2022} in all metrics.
\paragraph{Indoor experiments} 
To further assess model generalisation, we conduct additional experiments in an indoor domain transfer setting. Specifically, we benchmark our approach against a baseline experiment 'all-synthetic', trained only on supervised source data, and T2Net~\citep{zheng_t2net_2018}. However, as our experimental protocol evaluates a novel SceneNet-to-NYUv2 domain shift scenario not explored in the original work, it was necessary to retrain T2Net's framework for direct comparison. 
As shown in Table~\ref{tab:scenenet}, our approach significantly outperforms training the 'all-synthetic' baseline, trained without leveraging unlabelled NYUv2 data. Remarkably, we achieve gains over the prior work \citep{zheng_t2net_2018} across most metrics. We outperform \citet{zheng_t2net_2018}'s approach in every metric except the Absolute Relative Error.

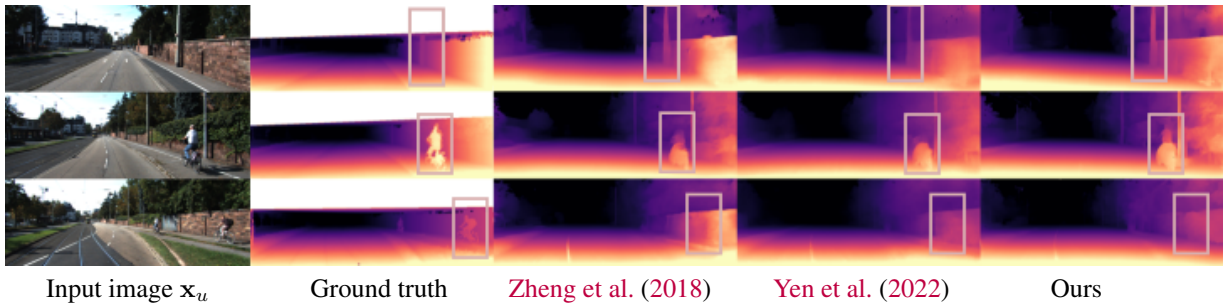
\begin{figure*}[h]
  \centering
   \includegraphics[width=0.98\linewidth]{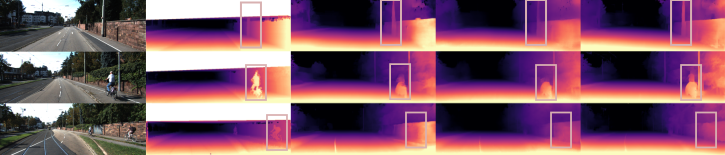}
   \put(-445,-12){Input image $\mathbf{x}_u$}
   \put(-345,-12){Ground truth}
   \put(-270,-12){\cite{zheng_t2net_2018}}
   \put(-170,-12){\cite{yen_3d-pl_2022}}
   \put(-65,-12){Ours}

   \caption{Qualitative results on KITTI~\citep{geiger_vision_2013_kitti} with models trained on vKITTI-KITTI. Ground truth depth is linearly interpolated for visualization.}
   \label{fig:kittiresults}
\end{figure*}
\paragraph{Qualitative results} In this setting, we compare our approach trained on the vKITTI-KITTI domain adaptation task and test it on the KITTI Eigen split~\citep{eigen_depth_2014} with previous state-of-the-art methods~\citep{yen_3d-pl_2022,zheng_t2net_2018}. Fig. \ref{fig:kittiresults} shows that  our approach better recovers fine-grained depth details in several examples. In particular, in the upper row our approach produces more accurate results for the lamp post. Similarly, in the second row our approach is capable of detecting the cyclist entirely, compared to 3D-PL.
{
\paragraph{Limitations} A key limitation of our approach is its reliance on synthetic depth maps possessing well-defined edges in the source domain. Due to the discontinuities at object boundaries present in these rendered depth maps, they differ from real-world captured depth maps which typically exhibit smoother transitions at edges.
The edge characteristics between our synthetic source depth maps and target domain data, which lacks such discontinuities, introduces a domain shift that our method seeks to overcome via target-domain consistency regularisation and source-based regularisation. A key assumption of our approach is therefore that the edge characteristics of depth maps in the target and source domains differ sufficiently. \\\\
Another key limitation of the proposed approach stems from its computational demands~(see also Tab.~\ref{tab:req}). The framework enforces consistency across multiple augmented views of target images as well as a supervised pairwise loss in the source domain. This joint optimisation comes at the cost of increased computational complexity. Scaling the framework to high-resolution imagery or large datasets would require more expensive GPU hardware.
}
\subsection{Ablation studies}
\label{sec:ablations}
Next, we analyse several key hyper-parameters involved in our proposed methodology. First, we explore the effect of varying the ratio parameter $r$.

Next, we evaluate different data augmentation techniques to understand their impact on our model's ability when no source domain samples are provided.
\paragraph{Importance of the ratio $r$}\label{parag:ablationr} The balance between unlabelled target consistency training samples and labelled source supervision samples is determined by the ratio $r$. To evaluate the importance of this hyperparameter, we conduct an ablation study examining different constant sampling ratios. 
The quantitative results presented in Table~\ref{tab:valuesr} indicate the sensitivity of our approach to the ratio hyperparameter $r$. Peak performance was observed for a ratio of $r = 2/1$, where twice as many supervised source images as unsupervised target images were sampled within each batch. Hyperparameters were chosen based on the obtained uncertainty $u$ (For a more detailed explanation please refer to \ref{ap:hypersel}).
Additionally, the effect of the ratio $r$ on the level of regularisation can be observed. A ratio $r \approx 1$ leads to weaker regularisation, as evidenced by the higher absolute and squared relative errors achieved with lower $r$ values. This is understandable, as a lower $r$ places less importance on the paired source loss term during training, therefore providing less regularisation pressure.
\begin{table*}[h]
\centering
\caption{Ablation study of our approach with ratio hyperparameter $r$. All approaches are capped to 50m.}
\label{tab:valuesr}
\footnotesize
\begin{tabular}{ccccccccccccc}
\hline
ratio $r$ &    AbsRel$\downarrow$ & SqRel$\downarrow$ &RMSE$\downarrow$& RMSE log $\downarrow$& A1$\uparrow$ &A2$\uparrow$&A3$\uparrow$ & Uncertainty $u$\\
\hline
${7}/{4}$ &  0.170 & 1.092 & \textbf{4.406} & 0.241 & 0.783 & 0.918 & 0.967 & 216.83 \\
${9}/{2}$ & 0.169 & 1.115 & 4.437 & 0.240 & 0.783 & 0.918 & 0.967 & 198.83\\
${6}/{5}$  &  0.183 & 1.441 & 4.635 & 0.252 & 0.775 & 0.912 & 0.963 & 219.29\\
${8}/{15}$ & 0.176 & 1.153 & 4.414 & 0.243 & 0.779 & 0.915 & 0.966 & 210.15\\
$2/1$ & \textbf{0.161} & \textbf{1.028} & 4.449 & \textbf{0.237} & \textbf{0.785} & \textbf{0.918} & \textbf{0.968} & \textbf{170.73}\\
\hline
\end{tabular}
\end{table*}

\paragraph{Comparison with different loss functions} 
Next, we conducted an additional experimental comparison using the virtual KITTI (vKITTI)~\citep{gaidon_vkitti_2016} and real KITTI~\citep{geiger_vision_2013_kitti} datasets~(see Tab.~\ref{tab:ablation_fm}). 
We compare our method to a standard baseline inspired by FixMatch~\citep{sohn_fixmatch_2020} that only leverages data augmentations within the target KITTI domain, i.e., without considering our multi-stream perturbation consistency losses. {Note: all experiments in Tab.~\ref{tab:ablation_fm} are conducted without using the pairwise source loss $\mathcal{L}_{\sum}^s$ except from the experiment 'Ours' in the last row}. In particular, we evaluate a setup using strong color-space augmentations only, and compare to RandAugment's geometric transformations. No weak augmentations were applied in all these setups.

The ablation study allowed us to assess the benefits of including geometric augmentations into our approach. Our experimental findings indicate that semi-supervised techniques for depth estimation consistently outperform direct pretraining~(see the "all synthetic" experiment in Tab.~\ref{tab:kitti50m}) without additional adaptation or regularisation. However, incorporating geometric data augmentations into the consistency regulariser provides a substantial boost in performance. Overall, our complete approach shows considerable improvements on the absolute relative error and squared relative error - two key metrics for assessing depth estimation accuracy.

\begin{table*}[h]
\centering
\caption{Ablation study: comparison with FixMatch~\citep{sohn_fixmatch_2020} baselines. All approaches are capped to 50m {and all experiments are subjected to pre-training.}}
\label{tab:ablation_fm}
\footnotesize
\begin{tabular}{ccccccccccccc}
\hline
Method &    AbsRel$\downarrow$ & SqRel$\downarrow$ &RMSE$\downarrow$& RMSE log $\downarrow$& A1$\uparrow$ &A2$\uparrow$&A3$\uparrow$ \\
\hline
Colorspace perturbations in $D_T$  & 0.183 & 1.393 & 4.697 & 0.250 & 0.775 & 0.912 & 0.964 \\
RandAugment perturbations in $D_T$  &  0.172 & 1.159 & 4.475 & 0.239 & 0.781 & 0.916 & 0.965 \\\hline
Ours {with $\mathcal{L}_{\sum}^s$ }& \textbf{0.161} & \textbf{1.028} & \textbf{4.449} & \textbf{0.237} & \textbf{0.785} & \textbf{0.918} & \textbf{0.968}\\
\hline
\end{tabular}
\end{table*}
{
\paragraph{Significance of combining $\mathcal{L}_{\sum}^s$ and $\mathcal{L}_{cons}^t$} To determine the effectiveness of our proposed approach combining source loss $\mathcal{L}_{\sum}^s$ and target loss $\mathcal{L}_{cons}^t$ we conduct an ablation study using the losses separately~(see Tab.~\ref{tab:comb}).}
\begin{table*}[h]
\centering
\caption{Comparison of our approach when combining and separating both the source loss $\mathcal{L}_{\sum}^s$ and the target loss $\mathcal{L}_{cons}^t$.}
\label{tab:comb}
\footnotesize
\begin{tabular}{cccccccccccccc}
\hline
Losses & AbsRel$\downarrow$ & SqRel$\downarrow$ &RMSE$\downarrow$ & RMSE log$\downarrow$ & A1$\uparrow$&A2$\uparrow$&A3$\uparrow$ \\
\hline
with $\mathcal{L}_{\sum}^s$, $\mathcal{L}_{cons}^t$&\textbf{0.161}&\textbf{1.028}&\textbf{4.449}&\textbf{0.237}&\textbf{0.785}&\textbf{0.918}&\textbf{0.968}\\
with $\mathcal{L}_{cons}^t$&0.171&1.178&4.544&0.248&0.781&0.912&0.965\\
with $\mathcal{L}_{\sum}^s$& 0.178 & 1.356 &5.040&0.258&0.779&0.912&0.961 \\
Only pretraining&0.173&1.413&5.062&0.264&0.767&0.906&0.957\\

\hline
\end{tabular}
\end{table*}{\\
These results reveal that the primary driver of the model's enhanced performance is the unsupervised consistency loss enforced on target perturbations (i.e. $\mathcal{L}_{cons}^t$). Specifically, the pairwise loss on the source domain (i.e. $\mathcal{L}_{\sum}^s$) alone provides only minimal regularisation benefits compared to the baseline. However, when combining both proposed objectives, we observe more noticeable gains according to certain error metrics such as absolute relative error (Abs Rel) and root mean squared error of the logarithm (RMSE log). This suggests that jointly leveraging source supervision together with unsupervised consistency across target transformations produces a regularisation effect superior to utilising either loss separately.

\section{Conclusion}

We formalised the problem of unsupervised domain adaptation for monocular depth estimation as consistency-based semi-supervised learning. Our key contributions are the simplicity and effectiveness of our method for training a single model to produce high quality depth estimates in the unlabelled target domain, based on the consistency of the predictions under different input augmentations. The proposed joint optimisation of the source and target domains with the proposed perturbation consistency loss and the pairwise source loss functions simplifies the training process compared to previous work that relies on complex training protocols and uses more than a single model. In our experiments, we showed promising results on the KITTI and NYUv2 datasets, which are considered standard for the domain adaptation task in monocular depth estimation.
\section{Acknowledgements}
Part of the research leading to these results is funded by the German Research Foundation (DFG) within the project 458972748. The authors would like to thank the foundation for the successful cooperation.

Additionally the authors gratefully acknowledge the scientific support and HPC resources provided by the Erlangen National High Performance Computing Center (NHR@FAU) of the Friedrich-Alexander-Universität Erlangen-Nürnberg (FAU). The hardware is funded by the German Research Foundation (DFG).

\bibliography{collas2024_conference}
\bibliographystyle{collas2024_conference}

\pagebreak
\appendix
\section{Appendix}
\subsection{Metrics}
We use the common evaluation metrics used for several depth estimation benchmarks and established by \cite{eigen_depth_2014}. These include:
\begin{itemize}
\item Root Mean Squared Error (RMSE):
\begin{equation}
RMSE = \sqrt{\frac{1}{N}\sum_{i=1}^{N}||d_i - \hat{d}_i||^2}
\end{equation}

\item RMSE log:
\begin{equation}
RMSE_{log} = \sqrt{\frac{1}{N}\sum_{i=1}^{N}||\log{d_i} - \log{\hat{d}_i}||^2}
\end{equation}

\item Squared Relative Error (Sq Rel):
\begin{equation}
Sq Rel = \frac{1}{N}\sum_{i=1}^{N}\frac{||d_i - \hat{d}_i||^2}{d_i^2}
\end{equation}

\item Absolute Relative Error (Abs Rel):
\begin{equation}
AbsRel = \frac{1}{N}\sum_{i=1}^{N}\frac{|d_i - \hat{d}_i|}{d_i}
\end{equation}

\item Accuracy under threshold $\delta$:
\begin{equation}
Accuracy_{\delta} = \frac{1}{N}\sum_{i=1}^{N}I(|\hat{d}_i - d_i| < (d_i*\delta))
\end{equation}
\end{itemize}

Where $d_i$ is the ground truth depth, $\hat{d}_i$ is the estimated depth, $N$ is the total number of samples, and $I(\cdot)$ is an indicator function that returns 1 if the condition holds, 0 otherwise. In the experimental results $A1$ refers to $\delta < 1.25$,$A2$ to $\delta < 1.25^2$ and $A3$ to $\delta<1.25^3$.
\subsection{Pretraining}
During pretraining, we solely use source domain data $D_S$ to train the depth estimation model $f(\cdot)$.
The following sections explain hyperparameter adjustment for the proposed pretraining approach and provide more information on the used augmentations during pretraining.
\paragraph{CutMix Hyperparameter $\alpha$}
We use a hyperparameter of $\alpha = 0.5$ to balance the information between the original and augmented images during CutMix~\citep{yun_cutmix_2019}. This ratio was chosen so that the model would not be overwhelmed by drastic augmentations, as monocular depth estimation requires an understanding of the surrounding scene context. With $\alpha=0.5$, patches from the two input images are mixed approximately equally, providing additional cues about depth relationships while still retaining most of the original scene structure. 
\paragraph{Other augmentations}
In addition to CutMix augmentations during the pre-training, we also employ standard color and geometric data augmentation techniques. Color augmentations, including random jitter, are applied to expose the model to variations in colorspace. Furthermore, geometric transformations such as random rotations are utilised. This can be seen in Fig. \ref{fig:jitter}.
\begin{figure}[h]
  \centering
   \includegraphics[width=0.9\linewidth]{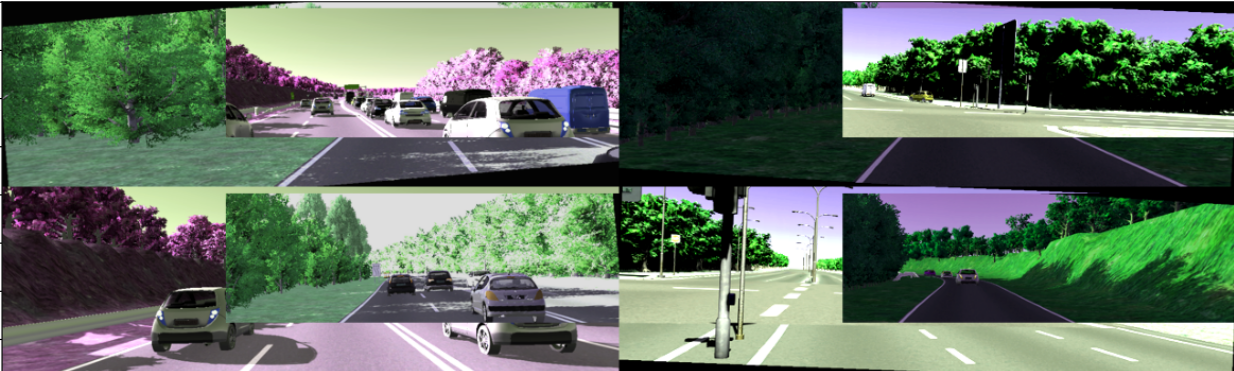}

   \caption{CutMix augmented images with random jitter and rotation augmentations in $D_S$ for images in vKITTI.}
   \label{fig:jitter}
\end{figure}
\begin{figure*}
  \centering
   \includegraphics[width=0.9\linewidth]{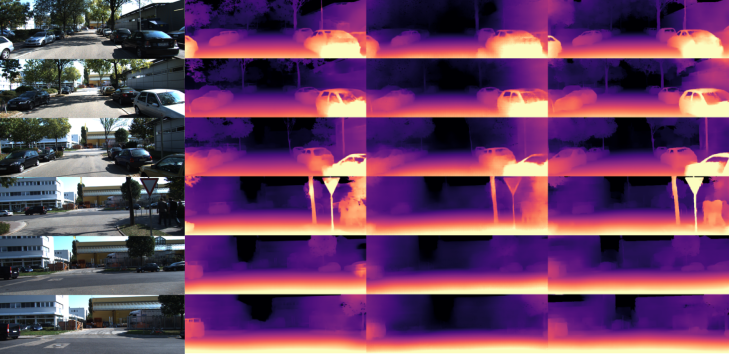}
   \put(-410,-12){Input image $\mathbf{x}_u$}
   \put(-295,-12){\cite{zheng_t2net_2018}}
   \put(-185,-12){\cite{yen_3d-pl_2022}}
   \put(-93,-12){$f(\cdot)$ after refinement}

   \caption{Additional qualitative results on KITTI eigen test samples $x_u$ between other state-of-the-art models.}
   \label{fig:addkittiresults}
\end{figure*}
\begin{figure*}
  \centering
   \includegraphics[width=0.9\linewidth]{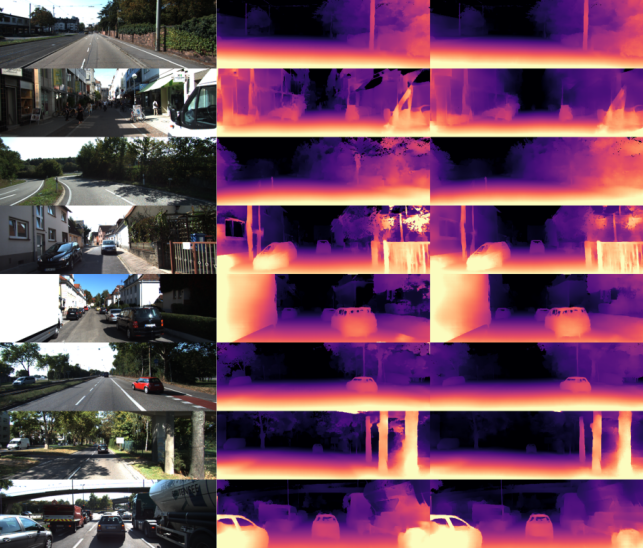}
   \put(-380,-12){Input image $\mathbf{x}_u$}
   \put(-240,-12){pretrained $f(\cdot$)}
   \put(-110,-12){$f(\cdot)$ after refinement}

   \caption{Comparison of depth predictions on KITTI eigen test samples $x_u$, after pretraining~(middle column) and after refinement~(right column).}
   \label{fig:pretrainvsfinal}
\end{figure*}

\paragraph{Quantitative results}
Our findings demonstrate that models pre-trained using the  augmentation pipeline incorporating CutMix, randomised colorspace jittering, and geometric rotations produce improved depth predictions compared to networks exclusively optimised on the labeled vKITTI source data without augmentations~(see Tab. \ref{tab:aug}). However, quantitative evaluation reveals the proposed approach does not achieve a performance similar to state-of-the-art methods for monocular depth estimation in unsupervised domain adaptation. 
\paragraph{Qualitative results}
Further qualitative results in the pretraining stage, where the model $f(\cdot)$ is only trained on heavily augmented source data in $D_S$, can be seen in Fig.~\ref{fig:pretrainvsfinal}. A comparison between the pretrained $f(\cdot)$ and $f(\cdot)$ after the proposed approach shows a clear smoothing effect of our approach. Sharp discontinuities present in the boundary regions of the pretrained $f(\cdot)$ maps are noticeably attenuated after refinement. This indicates the model $f(\cdot)$ learns to produce depth outputs that align more smoothly across pixel neighbours after leveraging both labelled source data and unlabelled target images during refinement.
\begin{table*}[h]
\centering
\caption{Depth estimation performance of our model $f(\cdot)$ after the pretraining in $D_S$. $f(\cdot)$ is trained on vKITTI only and tested on KITTI (50m cap). Different augmentation pipelines are compared.}
\label{tab:aug}
\footnotesize
\begin{tabular}{cccccccccccccc}
\hline
Augmentations & AbsRel$\downarrow$ & SqRel$\downarrow$ &RMSE$\downarrow$ & RMSE log$\downarrow$ & A1$\uparrow$&A2$\uparrow$&A3$\uparrow$ \\
\hline
None& 0.244 & 1.771 & 5.354 & 0.313 & 0.647 & 0.866 & 0.943 \\
CutMix, rotations, jitter&0.173&1.413&5.062&0.264&0.767&0.906&0.957\\
Rotations, jitter&0.187&1.538&5.104&0.266&0.756&0.901&0.958\\

\hline
\end{tabular}
\end{table*}

\paragraph{Training process} 
During refinement, both the source ($D_S$) and target ($D_T$) datasets are jointly used to update the monocular depth estimation model $f(\cdot)$. To augment the training data, we apply a combination of various techniques, which will be discussed in the following section. Subsequently, we provide more qualitative results. Finally, we present details on the computational resources required for the refinement process.
{
\subsection{Further ablation studies}
\paragraph{Number of perturbations in $\mathcal{L}_{cons}^t$}
To explore the effect of increasing target domain perturbations, we conduct an experiment varying the number of augmentation streams applied to target samples~(see Tab.~\ref{tab:losses}). Specifically, we evaluate model performance when employing two, three and four independent RandAugment~\citep{cubuk_randaugment_2019} augmentations of target samples simultaneously as multi-stream perturbations during training, in addition to the pairwise supervised source loss $\mathcal{L}_{\sum}^s$.
}
\begin{table*}[h]
\centering
\caption{Comparison of required GPU memory and performance metrics for two,three and four perturbations of target images for the target loss $\mathcal{L}_{cons}^t$.}
\label{tab:losses}
\footnotesize
\begin{tabular}{cccccccccccccc}
\hline
\# of perturbations & required GPU memory  & Abs Rel$\downarrow$ & A1$\uparrow$\\\hline
2&30GB&0.165&0.779\\
3&35GB&\textbf{0.161}&0.785\\
4&40GB&\textbf{0.161}&\textbf{0.786}\\
\hline
\end{tabular}
\end{table*}
{\\
It should be noted that we calculate the number of perturbations by counting each augmented sample alongside its corresponding original, non-augmented counterpart. For example, to obtain a total of three perturbations, we employ one baseline image paired with two transformed views.
Our experiments revealed that the quantity of perturbations applied to target domain data has a substantial impact on the efficacy of the proposed approach. While gains were observed for all perturbation counts investigated, performance improvements became more pronounced with more than two augmented samples. Peak results were attained with three perturbations, as evidenced by our key evaluation metrics.
As incorporating additional perturbations linearly increases computational overhead and memory requirements, we opted to use the three-perturbation setting due to its trade-off between achieved accuracy and demanded resources. }

\paragraph{Training time in comparison to other approaches} To facilitate quantitative comparison of computational requirements, we conducted an ablation analysis reporting training time and number of models for our method and relevant baselines. Since training duration scales with the number of models optimized as well as their architectures, we record both the total time spent training and the additional models employed beyond our approach. Across all experiments, we log two performance metrics(see Tab.~\ref{tab:req}).
}

\begin{table*}[h]
\centering
\caption{Comparison of the proposed approach with other methods in terms of the total training time, the number of additionally trained models and two performance metrics.}
\label{tab:req}
\footnotesize
\begin{tabular}{cccccccccccccc}
\hline
Method & Total training time & \# of trainable models &Abs Rel$\downarrow$ & A2$\uparrow$\\
\hline
Ours & $\approx$6h &1&0.161&0.918\\
~\cite{zheng_t2net_2018}&$\approx$8h&3&0.168&0.912\\
~\cite{yen_3d-pl_2022}&-&2&0.162&0.916\\
~\cite{lopez-rodriguez_desc_2020}&$\approx$18h&3&0.161&0.915\\
\hline
\end{tabular}
\end{table*}
{
Due to the unavailability of training code from 3D-PL~\citep{yen_3d-pl_2022}, we could not directly replicate their model optimization procedure and thus cannot report their total training time. While DESC~\citep{lopez-rodriguez_desc_2020} attains markedly better results than other referenced techniques, it employs a multi-stage pipeline incorporating additional models for object detection, sim-to-real translation, and depth estimation. Remarkably, our proposed method outperforms all prior work across standard evaluation metrics for this task while retaining a highly efficient single-model formulation.
\paragraph{Significance of $\mathcal{L}_{\sum}^s$} We conduct additional ablation experiments to further evaluate the impact of our proposed pairwise source loss $\mathcal{L}_{\sum}^s$~(see Tab.~\ref{tab:slosses}). First, we compare the standard per-sample $\mathcal{L}_1$ loss~$(\mathcal{L}_1^s(\hat{y}_{s,1},y_{s,1}))$ against our pairwise formulation. Secondly, we independently apply the pairwise loss~($\mathcal{L}_{\sum,sep}^s = \mathcal{L}_1(\hat{y}_{s,1},y_{s,1}) + \mathcal{L}_1(\hat{y}_{s,2},y_{s,2}))$ without jointly optimising the samples.\\
All experiments utilise three target domain perturbations for consistency regularisation, to isolate the effects of varying only the source supervision approach.
}
\begin{table*}[h]
\centering
\caption{Comparison of different implementations for the source loss.}
\label{tab:slosses}
\footnotesize
\begin{tabular}{cccccccccccccc}
\hline
source loss & AbsRel$\downarrow$ & SqRel$\downarrow$ &RMSE$\downarrow$ & RMSE log$\downarrow$ & A1$\uparrow$&A2$\uparrow$&A3$\uparrow$ \\\hline
no source loss&0.171&1.178&4.544&0.248&0.781&0.912&0.965\\
$\mathcal{L}_{\sum}^s$   & 0.161   & 1.028  & 4.449 & 0.237    & 0.785 & 0.918 & 0.968 \\
$\mathcal{L}_1^s$        & 0.167   & 1.143  & 4.468 & 0.241    & 0.783 & 0.913 & 0.965 \\
$\mathcal{L}_{\sum,sep}^s$ & 0.164   & 1.109  & 4.491 & 0.241    & 0.781 & 0.912 & 0.966 \\
\hline
\end{tabular}
\end{table*}
{\\
Across all experiments utilising the source loss $\mathcal{L}^s$, we observed a regularising effect relative to omitting source supervision entirely. However, upon closer inspection, the Abs Rel and Sq Rel error metrics revealed a notable performance gap. Specifically, the pairwise loss ${\mathcal{L}_{\sum}}^s$ substantially outperformed the standard per-sample ${\mathcal{L}1}^s$ and independently applied pairwise ${\mathcal{L}_{\sum,sep}}^s$ variants.
}
{
\paragraph{Hyperparameter selection}
\label{ap:hypersel}
Given the lack of depth annotations in the target domain, we required an unsupervised method to assess prediction performance without ground truth labels. Gradient-based uncertainty estimation~\citep{hornauer_gradient-based_2022} allows for such an evaluation under this unlabelled setting.\\
By approximating a model's uncertainty through perturbations and gradients, this technique enables quantification of a prediction's ambiguity even without a labelled target. As our goal was to validate the quality and reliability of our estimated depth maps for the unlabelled target domain, we opted to apply this approach.\\
The hyperparameters~($r$, learning rate, number of epochs) were optimised via predictive uncertainty estimation of the trained models. Models trained with various hyperparameter values were evaluated, model selection was then based on minimizing average estimated uncertainty.\\
Specifically, we generated reference depth predictions by augmenting inputs through horizontal flipping~\citep{hornauer_gradient-based_2022}. To approximate uncertainty, gradients of the convolutional blocks within the UNet-based depth estimation network, particularly in the expansion phase, were extracted.\\
 We performed gradient-based uncertainty analysis on the KITTI Eigen validation split comprising 888 to evaluate outdoor performance. For indoors, we applied the uncertainty estimation to a held-out 1000 image NYUv2 subset, both unseen during training. }

\subsection{Training process in $D_T$}
To gain deeper insights, ablation experiments are also conducted investigating the impact of perturbations applied via RandAugment~\citep{cubuk_randaugment_2019} mutli-stream consistency regularisation. RandAugment utilises randomly composed augmentations parameterised by the augmentation set $S$, depth $n$ and severity $m$. Specifically, we analyse variations in these hyperparameters which control the RandAugment transformations enforcing predictive consistency between augmented unlabelled target samples. The parameters studied include the augmentation set $S$ from which augmentations are selected, as well as the ranges $n$ and $m$ defining the permissible number and strength of augmentations in a random augmentation chain. Our default augmentation set is similar to the configuration presented in the FixMatch~\citep{sohn_fixmatch_2020} literature, though we additionally evaluate removing geometric transforms to analyse their contribution. The official augmentation set $s_{fm}$ used in FixMatch are: \begin{verbatim}
s_fm = [AutoContrast(), 
    Brightness(), 
    Color(), 
    Contrast(), 
    Equalize(), 
    Identity(), 
    Posterize(), 
    Rotate(), 
    Sharpness(), 
    ShearX(),
    ShearY(), 
    Solarize(), 
    TranslateX(), 
    TranslateY()]
\end{verbatim}
In our ablation studies we removed all geometric augmentation, such that $s_{geo}$:
\begin{verbatim}
s_geo = [AutoContrast(), 
    Brightness(), 
    Color(), 
    Contrast(), 
    Equalize(), 
    Identity(), 
    Posterize(), 
    Sharpness(), 
    Solarize()]
\end{verbatim}
In Tab.~\ref{tab:svssgep} the performance of the depth estimation model $f(\cdot)$ is compared in mulitple variations of the hyperparameters $s$,$n$ and $m$. The results indicate that the augmentation set $s_{geo}$ is outperformed by $s_{fm}$. The best performance is given for the parameter combination $s=s_{fm}$,$n=1$ and $m=7$ although performance gaps in general are minor for parameters within $s_{fm}$. 
During our analysis of the augmented data, we observed that for the parameter setting of the depth parameter $n=1$, the transformations sometimes did not perturb the input at all,  limiting their regularisation effect. To address this, we introduce a static CutOut augmentation applied to the input images prior to the other transformations.
\begin{table*}[t]
\centering
\caption{Comparison of the refinement process trained on vKITTI-KITTI and tested on the KITTI eigen split (capped to 50m), depending on the augmentation set $s$, depth $n$ and severity $m$ of the augmentation in $D_T$.}
\label{tab:svssgep}
\footnotesize
\begin{tabular}{cccccccccccccc}
\hline
$s$ & $n$ & $m$ & AbsRel$\downarrow$ & SqRel$\downarrow$ &RMSE$\downarrow$ & RMSE log$\downarrow$ & A1$\uparrow$&A2$\uparrow$&A3$\uparrow$ \\
\hline
$s_{geo}$ & 1 & 5&0.172&1.164&4.490&0.244&0.781&0.916&0.966\\
$s_{fm}$& 1& 5& 0.169&1.087&\textbf{4.428}&0.239&0.783&0.917&0.967\\
$s_{fm}$& 5&7&0.173&1.123&4.41&0.242&0.781&0.916&0.966\\
$s_{fm}$& 1&7&\textbf{0.161}&\textbf{1.028}&4.449&\textbf{0.237}&\textbf{0.785}&\textbf{0.918}&\textbf{0.968}\\
\hline
\end{tabular}
\end{table*}
\paragraph{Qualitative results}
To further evaluate our approach on more complex KITTI  scenes, Fig.~\ref{fig:addkittiresults} compares our refined model $f(\cdot)$ to other state-of-the-art methods using denser samples from the KITTI eigen test split. Qualitatively, our method produces depth maps that better preserve fine structural details throughout the scenes.

\begin{figure*}
  \centering
   \includegraphics[width=0.9\linewidth]{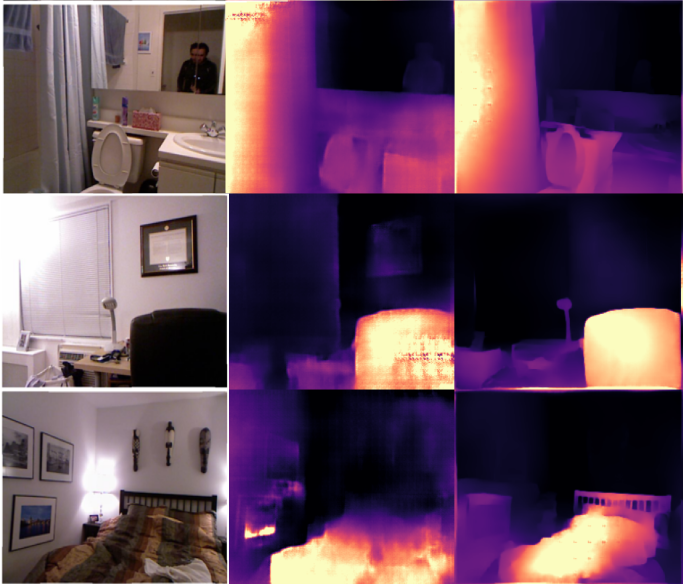}
   \put(-390,-12){Input image $\mathbf{x}_u$}
   \put(-245,-12){\cite{zheng_t2net_2018}}
   \put(-120,-12){$f(\cdot)$ after refinement}

   \caption{Qualitative results on NYUv2 eigen test samples $x_u$ to  another state-of-the-art model. Both models were trained in the indoor domain adaptation setting~(SceneNet-NYUv2) and are evaluated on NYUv2 eigen test samples.}
   \label{fig:qualnyu}
\end{figure*}
Prior work tends to oversmooth object boundaries and thin structures such as street signs. In contrast, $f(\cdot)$ trained with our domain adaptation framework more sharply smoothes objects at boundaries while still maintaining consistent predictions. This indicates its capability to balance high-frequency content with regularizing across image domains $D_S$ and $D_T$.

To further demonstrate the effectiveness of our approach in indoor scenes, we present qualitative evaluations on the NYUv2~\citep{nyu_depth} eigen test set~\citep{eigen_depth_2014} in Fig.~\ref{fig:qualnyu}. Similar to the outdoor KITTI results, our method preserves fine-scale structural details that prior methods often overlook.

For example, as shown in the first row of Fig.~\ref{fig:qualnyu}, our predicted depth maps are able to sharply identify objects like toilet seats with thin structures. Additionally, Fig.~\ref{fig:qualnyu} shows that our approach accurately infers the scene layout while also capturing small objects such as a desk lamp against complex indoor backgrounds.

\paragraph{Hardware requirements}
Empirically, we found a ratio parameter $r=2$ performed best across indoor and outdoor scenes. With a batch size $N$ of 12, the concatenated inputs contain $N_{concat}=42$ images as:
\begin{equation}
    \label{eq:pairlossbs}N_{concat}=\underbrace{3\times N\times\frac{1}{r}}_{[x_u,\tilde{x}_{u,1},\tilde{x}_{u,2}] \in D_T}+\underbrace{2\times N\times r}_{[x_{s,1},x_{s,2}] \in D_S}\bigg|_{r=2,b=12}
\end{equation}

For KITTI resized to 640x192, this requires 35GB of GPU memory. With NYUv2 at 256x192, memory usage is 20GB.

Higher resolutions increase memory cost due to redundant feature representations. We also tested varying $N$, but larger batches exceed GPU limits whereas smaller $N$ weakened the overall regularisation effect. Given hardware constraints, the above configuration provides the optimal trade-off between batch size, memory requirements, and regularisation effect during our refinement approach.

\end{document}